\documentclass[journal]{IEEEtran}

\ifCLASSINFOpdf

\else

\fi

\usepackage{bbding}
\usepackage{bbm}
\usepackage{algorithm}
\usepackage{algorithmic}
\usepackage{multirow}
\usepackage{subfigure}
\usepackage{amssymb}
\usepackage{amsmath}
\usepackage{mathrsfs}
\usepackage{bm}
\usepackage{graphicx}
\usepackage{lineno,hyperref}
\usepackage{epsfig}
\usepackage{rotating, multirow}
\usepackage{epstopdf}
\usepackage{float}
\usepackage{verbatim}
\usepackage{booktabs}
\usepackage{array}
\usepackage{enumitem}

\makeatletter

\newcommand{\Rmnum}[1]{\expandafter\@slowromancap\romannumeral #1@}
\makeatother

\begin{document}
	
	%
	\title{EVC-MF: End-to-end Video Captioning Network with Multi-scale Features}

	\author{Tian-Zi Niu, Zhen-Duo Chen, Xin Luo, Xin-Shun Xu, Senior Member, IEEE}

	\maketitle
	
	\begin{abstract}
		Conventional approaches for video captioning leverage a variety of offline-extracted features to generate captions. Despite the availability of various offline-feature-extractors that  offer diverse information from different perspectives, they have several limitations  due to fixed parameters. Concretely,   these extractors  are solely  pre-trained on image/video comprehension  tasks, making them less adaptable to video caption datasets. Additionally, most of these extractors only capture features prior to the classifier of the pre-training task, ignoring a significant amount of valuable shallow information. Furthermore, employing  multiple offline-features may introduce redundant information. To address these issues, we propose an end-to-end encoder-decoder-based network (EVC-MF) for video captioning, which efficiently utilizes multi-scale visual and textual features to generate video descriptions. Specifically, EVC-MF consists of three modules. Firstly, instead of  relying on multiple feature extractors, we directly feed   video frames  into a transformer-based network to obtain multi-scale visual features and update feature extractor parameters. Secondly, we fuse the multi-scale features and input them into a masked encoder to reduce redundancy and encourage learning useful features.  Finally, we utilize an enhanced transformer-based decoder, which can efficiently  leverage shallow textual  information, to generate video descriptions. To evaluate our proposed model, we conduct extensive experiments on benchmark datasets. The results demonstrate that EVC-MF yields competitive performance compared with the state-of-the-art methods.
		
	\end{abstract}
	

	
	\begin{IEEEkeywords}
		Video captioning, Multi-scale Features, End-to-end Network.
	\end{IEEEkeywords}
	
	%
	\IEEEpeerreviewmaketitle

	\section{Introduction}\label{intro}
	\IEEEPARstart{D}eveloping conversational systems that can both reliably comprehend  the world and effortlessly interact with humans is  one of the long-term goals of  artificial intelligence community. An dynamic  and thriving benchmark in this field is video captioning, integrating  research in visual understanding and natural language processing. Specifically, it entails automatically generating a semantically accurate description for a given video.  Despite recent promising achievements in this area, it remains a challenging task due to two primary reasons: 1) videos encompass intricate  spatial and temporal information compared to images; 2) there exists  an inherent gap between visual and natural language,  as their fundamental  syntax for conveying information differs significantly.
	
\begin{figure}
		\subfigure[Prior Works]
		{\begin{minipage}{0.48\linewidth}
				\centering	
				\includegraphics[height=4.5cm]{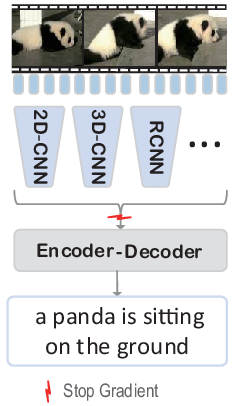}
				\label{fig_tran}
		\end{minipage}}
		\subfigure[EVC-MF]
		{\begin{minipage}{0.48\linewidth}
				\centering	
				\includegraphics[height=4.5cm]{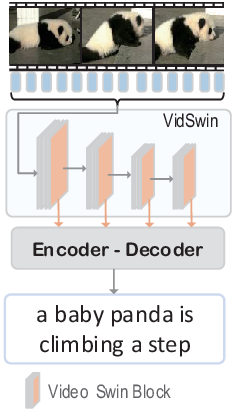}
				\label{fig_our}
			\end{minipage}
		}
		\caption{Comparison between previous works and EVC-MF.}
		\label{fig_mul_layer}
	\end{figure}

	Inspired by machine translation, most recent visual captioning methods have adopted the encoder-decoder framework \cite{Zhang_2021_IC1,Zhang_2021_DEN,Wang_2022_CMG}. Naturally, some of them focus on designing a suitable encoder to learn more efficient video representation. For example, early approaches \cite{Yao_2015_TS,Song_2017_HA} typically  employ a pre-trained convolutional neural network (CNN) as an encoder to extract appearance  features. However, relying solely on appearance features makes it challenging to fully represent all contents within a video. Consequently, researchers have successively incorporated additional  information, such as action features \cite{Yao_2015_TS,Yu_2016_HRNN}, object features\cite{Pan_2020_STG,zhang_2020_ORG,Yan_2020_STAT}, and trajectory features \cite{Hua_2022_ARL}, to the encoder (Fig \ref{fig_tran}). Furthermore, there are also some feature fusion-based encoders proposed to effectively utilize these multi-modal features \cite{Dong_2023_SEG,Hori_2017_ABM}.

	 Although significant progress has been made with these methods, they typically rely on one or more offline-feature-extractors and encounter the following problems: 1) multiple extractors necessitate  higher  computational resources; 2) most offline-feature-extractors are pre-trained on tasks irrelevant to video captioning,  
	 making their adaptation to video captioning difficult; 3) most offline-feature-extractors only extract features before the classifier, disregarding  the valuable  information at shadow levels; 4) models employing offline-feature-extractors lack end-to-end training. 
	 
	 The other focus of encoder-decoder models lies in generating semantically correct and linguistically natural captions. Currently, caption generation is primarily  using autoregressive (AR) decoding, \textit{i.e.}, generating each word conditionally on previous outputs. For example, most methods utilize   recurrent neural networks, \textit{e.g.} GRU, LSTM,  with self-attention  mechanism or transformer as decoder. However, these methods treat the input sequence merely as a collection of tokens  and only independently calculate the attention weights between any two tokens within this collection. Consequently, they fail to consider the shallow textual information () when calculating dependencies among tokens.

\begin{figure*}
		\centering
		\includegraphics[width=13cm]{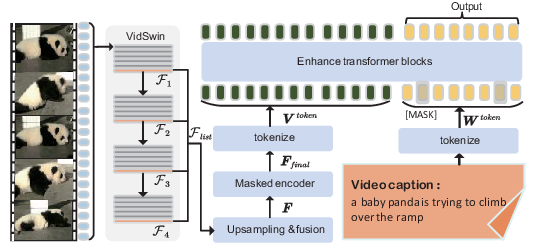}
		\caption{Illustration of the proposed framework, \textit{i.e.} EVC-MF.}
		\label{fig_framework}
	\end{figure*}

	To address the issues caused by the  offline-features and conventional decoders,  we propose a novel  \textbf{E}nd-to-end \textbf{V}ideo \textbf{C}aptioning network  with \textbf{M}ulti-scale \textbf{F}eatures, namely EVC-MF. The model comprises three modules to tackle these issues. Firstly, our objective  is to find a feature extractor that takes raw video frames as input and requires less computation. Therefore, as depicted in Fig. \ref{fig_our}, we adopt VidSwin \cite{Liu_2022_swin3D} to extract multi-scale visual features from raw frames as the initial video representation instead of relying on multiple offline-feature-extractors. Consequently, our feature extractor parameters can be fine-tuned based on the video caption dataset. Secondly, we feed the multi-scale visual features to a masked encoder to obtain a video tokens sequence. Specifically, we upsample the multi-scale visual features  to a uniform size and merge them into a feature map sequence. However,  this sequence contains redundant information. Notably, previous work \cite{He_2022_MA} has demonstrated  that masking a very high portion of random patches encourages learning useful features while  reducing  redundancy. Inspired by this, we segment  each feature map into multiple regions of varying sizes and randomly mask one region of each frame to derive  the final video representation.  Finally, we input the video representation into  an enhanced transformer-based decoder to generate  semantically accurate  captions. To make full use of shallow textual  information, we convert internal states of different   layers into global contextual information. Thus, the shallow textual  features are utilized  for computing  the correlation  between elements. Additionally, EVC-MF is trained with an end-to-end manner. The overall  structure of our model is illustrated  in Fig. \ref{fig_framework}. 
	
	Our main  contributions are summarized as follows:
	\begin{itemize}[leftmargin=*]
		\item  We propose  a novel end-to-end encoder-decoder-based network (EVC-MF) for video captioning to efficiently utilize multi-scale visual features and textual  information.
		\item  We design a masked encoder to integrate features of different sizes, promoting the  learning useful information while reducing  redundancy.
		\item   We propose an enhanced transformer-based decoder that effectively leverages shallow textual  information to  generate semantically correct captions.
		\item We conduct extensive experiments on benchmark datasets to  demonstrate the effectiveness of our method, and our model achieves competitive performance.
	\end{itemize}

	The rest of this paper is organized as follows. Section \ref{section:related} provides an overview of related works. Section \ref{section:our} elaborates on our proposed model, including the feature extractor, the masked encoder and the transformer-based decoder. Section \ref{section:experiment} presents  experimental results and  in-depth analyses, followed by the conclusion in Section \ref{section:conclusion}.
	
	\section{Related Work}\label{section:related}
	Pioneering models for the visual captioning task are mainly template-based methods \cite{Pradipto_2013_template1,Jesse_2014_template3,Socher_2010_tem_im,Kulkarni_2013_tem_im}, which employ predefined grammar rules and manual visual features to generate fixed descriptions.  However, these approaches are  significantly constrained by the predetermined templates, making them difficult to generate flexible and satisfactory descriptions. 
	
	 With the advancement of deep learning, sequence learning methods  gradually supplanting  template-based approaches to emerge as the prevailing paradigm for visual captioning.  Generally, a sequence learning method usually adopts an encoder-decoder framework  to convert visual information into textual information. Recently, several  state-of-the-art methods have proposed novel  encoder schemes, while others have made improvements on the decoder. In the following subsections, we comprehensively review these advancements from both encoder and decoder perspectives.
	
	\subsection{Encoder-design Methods}
	A high quality encoder should  encode visual contents into discriminative features that can be easily processed by machines. Currently, due to the limitations of computer computing power and model computation volume, most encoders  employ different offline extractors to obtain multi-modal visual representations.
	For example, PickNet \cite{Chen_2018_PickNet} employs the output of the final convolutional layer of ResNet-152 as the video representation for video captioning task. Similarly, Wang \textit{et al.} \cite{Bairui_2018_RecN} proposed RecNet, which  utilizes Inception-V4 pre-trained on  ILSVRC2012-CLS classification dataset for offline feature extraction. However, relying solely on a single CNN feature extraction may  lead to overlooking crucial information.
	
	By extracting features from multiple perspectives, a model can acquire  a more comprehensive understanding of the video. Consequently, researchers have progressively incorporated  diverse  perspectives information to the encoder. For instance,  to capture temporal information, numerous video captioning methods recommend using offline action features to enhance video comprehension. Specifically, in addition to employing 2D features, MARN \cite{Pei_2019_MARN} integrates  offline optical flow  to obtain a more accurate video representation. In addition, MGSA \cite{Chen_2019_MGSA}, POS-CG \cite{Wang_2019_POSCG}, and CANet \cite{Song_2023_CANet} employ  pre-trained 3D-CNN models, such as C3D, I3D, and 3D-ResNeXt respectively, to extract offline action information for video captioning. More recently, it is verified that incorporating more detailed features is beneficial  for  characterizing  the semantic of  images or videos. For example, STG-KD \cite{Pan_2020_STG} for video captioning and Up-Down \cite{Anderson_2018_UD} for image captioning demonstrate that object features and their interactions facilitate generating  detailed visual descriptions. Hua \textit{et al.} \cite{Hua_2022_ARL} showed that trajectory-based feature representation contributes significantly to video  captioning. Furthermore,  several approaches \cite{Dong_2023_SEG,Chen_2020_PMI} use feature fusion to enhance visual understanding. 
	
	However, as mentioned previously, there exist  certain limitations that hinder further advancements in visual captioning using offline features. The primary issue is that the  parameters of these offline feature extractors  are exclusively  pre-trained  for image/video comprehension tasks, posing difficulties in their adaptation to different video captioning datasets.  Accordingly, the end-to-end approaches are initially  applied in image captioning \cite{Fang_2022_E2Ei,Wang_2022_E2Ei}. 
	Given that videos encompass  more information and complex content than images, there are still many problems to be solved in end-to-end video captioning.  For instance, it is  difficult to capture and analyze contextual scenes and track  objects  movements  throughout a video. Additionally, videos often possess a high temporal dimension which can significantly increase model complexity. Training such models necessitates substantial hardware resources and time investment, potentially rendering them impractical for real-world applications.
	\subsection{Decoder-design Methods}
	A successful decoder should generate a semantically correct description using the above visual representation. Currently, most approaches adhere to either  autoregressive (AR) decoding or non-autoregressive (NA) decoding methods. 
	Autoregressive decoder generates sentences word by word, with each word conditional on the previously generated output. For instance, some methods \cite{Li_2015_ATT,Lianli_ABSC_2017} draw inspiration from machine translation task and employ single- or multi-layer LSTMs as decoders. Though different but along this line, another variant of RNN GRU is also commonly used as a decoder for visual captioning \cite{Karpathy_2015_G1,Aafaq_2019_G2}. Additionally, SeqVLAD  \cite{Xu_2018_VLAD}  and ConvLSTM \cite{Shi_2015_ConvLSTM} suggest using convolutional recurrent neural networks  as decoder to integrate the advantages of RNN and CNN. MS-RNN  \cite{song_2019_msrnn}, which comprises a multimodal LSTM (M-LSTM) layer and a novel backward stochastic LSTM (S-LSTM) mechanism, recommends considering subjective judgments and model uncertainties to improve video captioning performance. 
	
	More recently, with the remarkable progress of self-attention mechanism in various domains, transformer-based decoders have garnered  increasing attention. For instance, Lin \textit{et al.} \cite{Lin_2022_swinbert} proposed a transformer-based decoder with sparse attention, which can avoid the inherent redundancy in consecutive video frames. Additionally, Chen \textit{et al.} \cite{Chen_2018_TVT} introduced the Two-View Transformer (TVT), which includes two types of fusion blocks in decoder layers for combining different modalities effectively. 
	
	Furthermore, transformers are also employed as non-autoregressive decoders for parallel word generation to achieve significant inference speedup. For instance, Yang \textit{et al.} \cite{Yang_2021_NA} proposed a transformer-based non-autoregressive decoding model (NACF) to deal with slow inference speed and unsatisfied caption quality in video captioning. Similarly, O2NA \cite{Liu_2021_O2NA}, another transformer-based non-autoregressive decoding model, tackles the challenge of  controllable captioning by injecting strong control signals conditioned on selected objects, with the advantages of fast and fixed inference time. 
	
	In summary, transformer-based decoders have been increasingly successful in visual captioning tasks. However, most of them only focus on the pairwise  relationship between tokens independently, which may result in the neglect of  crucial shallow textual information.
	\section{Methodology}\label{section:our}
	\subsection{Overall Framework}
	The framework of EVC-MF is illustrated  in Fig. \ref{fig_framework}, comprising  a feature extractor, a masked encoder and  an enhanced transformer-based decoder. Specifically, we first uniformly sample $ T $ raw frames $ \{\bm{\mathcal{V}}_{t}\}_{t=1}^{T} $, each frame consists of  $ H\times W\times 3 $ pixels, \textit{i.e.} $ \bm{\mathcal{V}}_{t} \in \mathbb{R}^{H\times W\times 3} $. Then, we feed them into the feature extractor to extract grid features $ \mathcal{F}_{list}=\{\bm{\mathcal{F}}_{m}\}_{m=1}^{M} $ from each block of the extractor, where $ M $ denotes the number of blocks. Subsequently, we upsample each feature map $ \bm{\mathcal{F}}_{m}$ to a same size  and merge them into a feature  sequence $ \bm{F} \in  \mathbb{R}^{\frac{T}{2} \times \frac{H}{8}\times \frac{W}{8}\times C}$, where $ C $ is the channel dimension. We then present multiple regions with varying degrees of coarseness on each feature map of $ \bm{F} $. To encourage learning useful information and reduce
	redundancy, we randomly mask one region of each feature map in the sequence and  obtain the final video representation $ \bm{F}_{final} \in \mathbb{R}^{(\frac{T}{2}-1)\times \frac{H}{32}\times \frac{W}{32}\times C} $ through a 3D averaging pooling layer. Finally, we input $ \bm{F}_{final} $ to an enhanced transformer-based decoder to  generate a text sentence $ \hat{\mathcal{S}} = \{\hat{\boldsymbol{w}}_{1},\hat{\boldsymbol{w}}_{2},\dots,\hat{\boldsymbol{w}}_{N}\} $ containing $ N $ words to describe the video content. Further elaboration on each module is provided in the following subsections.
	\subsection{Feature Extractor}
	As mentioned previously,  most video captioning models using multiple extractors are  difficult to be trained end-to-end, thus  limiting their performance. Fortunately, VidSwin  \cite{Liu_2022_swin3D} achieves a favorable speed-accuracy trade-off and has made significant achievements in human action recognition. Therefore, we utilize  VidSwin as our feature extractor to encode raw video frames as multi-scale features. Concretely, we feed the raw video frames sequence
	$ \bm{V} \in  \mathbb{R}^{T\times H\times W\times 3}$ into VidSwin to extract grid features from each block, formally,
	\begin{equation}
		\label{eq_VidSwin}
		\begin{aligned}
			&\bm{\mathcal{F}}_{0}  = \textit{Be} (V),\\
			&\bm{\mathcal{F}}_{m} = \textit{Bl}_{m} (\bm{\mathcal{F}}_{m-1}),
		\end{aligned}
	\end{equation}
	where $ m $ is the number of the block in VidSwin, $ \textit{Be} $ and $ \textit{Bl}_{m} $ are the patch embedding layer  and the swin transformer block of VidSwin, respectively. Please refer to \cite{Liu_2022_swin3D} for more details about VidSwin. Subsequently,  we obtain a list of feature maps $ \mathcal{F}_{list} =\{\bm{\mathcal{F}}_{m}\}_{m=1}^{M} $, where $ \bm{\mathcal{F}}_{m} \in \mathbb{R}^{\frac{T}{2} \times \frac{H}{8 \times m }\times \frac{W}{8 \times m}\times C_{m} }$. 
	\subsection{Masked Encoder} 
	\begin{figure}
		\centering
		\includegraphics[width=4.5cm]{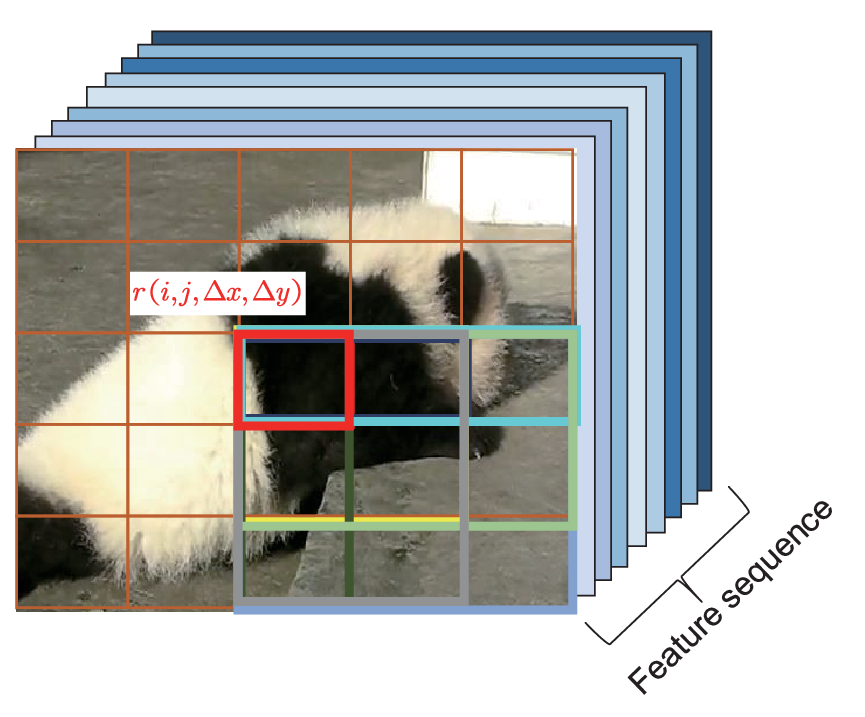}
		\caption{Illustration of $ R_{(i,j)} $ based on an andhor $ (i,j) $.}
		\label{fig_regions}
	\end{figure}	
	Obviously, the list $ \mathcal{F}_{list} $ contains  a substantial amount of redundant information. To integrate valuable information and reduce redundancy, we propose  a masked encoder. Specifically, we initially  feed    elements in  $ \mathcal{F}_{list} $  into a series of upsampling modules  to standardize their shapes. Each upsampling module contains a linear function $ \psi_{m} (\cdot) $ and an upsampling function $ \Psi_{m} (\cdot) $. The formulas are defined as follows,
	\begin{equation}
		\label{eq_upsample}
		\begin{aligned}
			&\bm{\Gamma}_{m} = \psi_{m} (\bm{\mathcal{F}}_{m}),\\
			&\tilde{\bm{\mathcal{F}}}_{m} = \Psi_{m} (\bm{\Gamma}_{m}),\\
			&\bm{F} =\left[\tilde{\bm{\mathcal{F}}}_{1},\tilde{\bm{\mathcal{F}}}_{2}, \cdots, \tilde{\bm{\mathcal{F}}}_{M}\right],
		\end{aligned}
	\end{equation}
	where $ \bm{\Gamma}_{m} $ is an intermediate variable, $ \tilde{\bm{\mathcal{F}}}_{m}  \in \mathbb{R}^{\frac{T}{2} \times \frac{H}{8}\times \frac{W}{8}\times \frac{C}{M} }$, $ \bm{F} \in  \mathbb{R}^{\frac{T}{2}\times \frac{H}{8}\times \frac{W}{8}\times C} $. 
	
	After that, to further process the features, we present multiple regions with varying level of coarseness on each feature map $ \bm{F}[t] $. The coarseness is determined by the size of a rectangular. As illustrated  in Fig. \ref{fig_regions}, we initially  divide  the feature map into $ \frac{H}{g} \times \frac{W}{g} $ grids, where each grid  has an area of  $ g \times g $. Then, we define the smallest region $ r(i,j, \Delta x,\Delta y) $ with  height $ \Delta y $ and  width $ \Delta x $ at  anchor point $ (i, j) $, \textit{i.e.} top-left corner grid point. Using $ r(i,j, \Delta x,\Delta y) $, we derive a set of regions 
	$ R_{(i,j)} = \{  r(i,j, w\Delta x, h\Delta y)| h,w \in \{1,2,\cdots\} , i+h \Delta y < \frac{H}{g}, j+w \Delta x < \frac{W}{g} , Ar(r)<\delta HW \} $ by changing their widths and heights, where $ Ar(r) $ denotes the area of the element, $ \delta $  is a threshold to ensure that the masked area does not exceed certain limits. Consequently, for different spatial locations $ (i,j) $, we can obtain different sets of rectangles $ R_{(i,j)} $. Ultimately, we obtain the set $ \mathcal{R} = \{R_{(i,j)} | 0<i<\frac{H}{g}, 0<j<\frac{W}{g} \}$ of regions with different coarseness of the whole feature map.  We randomly sample a sequence of regions $ \tilde{\mathcal{R}} = \{r_{1},r_{2},\cdots,r_{T}\} $, where $ r_{t} =r(i_{t},j_{t},w_{t}\Delta x,h_{t}\Delta y) $. After obtaining $ \tilde{\mathcal{R}} $, we can easily get the masked feature sequence $ \tilde{\bm{F}} $,
	\begin{equation}
		\label{eq_mf}
		\tilde{\bm{F}}[t][i][j]=
		\begin{cases} 
			\bm{F}[t][i][j],  & \mbox{if } (i,j) \not\in r_{t}\\
			\bm{0}_{C}, & \mbox{if } (i,j) \in r_{t},
		\end{cases}
	\end{equation}
	where $ \bm{0}_{C} $ is a C-dimensional zero vector. Finally, we feed $ \tilde{\bm{F}} $ to a 3D averaging pooling layer $ \rho (\cdot) $ to obtain the final video representation,
	\begin{equation}
		\label{eq_mp}
		\bm{F}_{final} = \rho (\tilde{\bm{F}}).
	\end{equation}
	
	\subsection{Enhanced Transformer-based Decoder} 
	Decoder aims to generate a semantically correct description based on the video representation. However, in most transformer-based decoders, the focus primarily lies on the individual relationships between two tokens, which may result in a loss of shallow textual  information. In this paper, we employ an enhanced transformer-based decoder to produce precise captions. Specifically, the input to the decoder is split into two parts: text tokens and visual tokens. Among them, the text tokens $ \bm{W}^{token} $ contain semantic and positional embedding, \textit{i.e.} $ \bm{W}^{emb} $ and $ \bm{P}^{emb} $, about the words in the caption, which is formulated as follows,
	\begin{equation}
		\label{eq_wordtoken}
		\begin{aligned}
			&\bm{W}^{emb}  = [\{\phi_{w} (\bm{w}_{n}^{e})\}_{n=1}^{N}], \\	
			&\bm{P}^{emb}  = [\{\phi_{p} (\bm{p}_{n}^{e})\}_{n=1}^{N}], \\	
			&\bm{W}^{token} = \bm{W}^{emb}+\bm{P}^{emb},\\
		\end{aligned}
	\end{equation}
	where $ \bm{W}^{token}, \bm{W}^{emb} $ and $ \bm{P}^{emb} \in \mathbb{R}^{N \times d} $; $ \left[ \dots \right] $ denotes concatenation; $ \phi_{w} $ and $ \phi_{p} $ are the embedding functions; $ \bm{w}_{n}^{e} $ and $ \bm{p}_{n}^{e} $ are the one-hot vectors of word $ \bm{w}_{n} $ and position $ n $, respectively.
	For the second one, we tokenize the video representation $ 	\bm{F}_{final} $ along the channel dimension and employ  a linear function  to ensure  dimensional  consistency with  $ \bm{W}^{token}$, 
	\begin{equation}
		\label{eq_visualtoken}
		\begin{aligned}
			&\bm{\Lambda}  = \varphi(\bm{F}_{final} ), \\		
			&\bm{V}^{token} = \psi_{v} (\bm{\Lambda}) ,\\
		\end{aligned}
	\end{equation}
	where $ \bm{\Lambda} \in \mathbb{R}^{[(\frac{T}{2}-1) \cdot \frac{H}{32} \cdot \frac{W}{32}]\times C}$ is an intermediate variable,  $ \psi_{v} (\cdot) $ is a linear function, $  \varphi(\cdot) $ denotes the tensor dimensional change function.
	Thus, we obtain  $ N $ text tokens and $(\frac{T}{2}-1) \cdot \frac{H}{32} \cdot \frac{W}{32} $ visual tokens. These tokens are combined to form the final input for the decoder $ \bm{I} = [\bm{W}^{token},\bm{V}^{token}] \in \mathbb{R}^{[N+(\frac{T}{2}-1) \cdot \frac{H}{32} \cdot \frac{W}{32}]\times d}$.

	
	As mentioned previously, our decoder is based on transformer. Upon receiving the input tokens, the traditional transformer based decoder \cite{Li_2020_Oscar,Hu_2021_VIVO} feeds them to a self-attention module with multiple layers to obtain the final output.
	A layer of the traditional transformer  is formulated as,
	\begin{equation}
		\label{eq_trtr}
		\begin{aligned}
			& \bm{Q}, \bm{K} , \bm{V}  = \bm{I}\bm{W}_{q}, \bm{I}\bm{W}_{k},\bm{I}\bm{W}_{v} \\		
			&\bm{H}=\Big(softmax(\frac{\bm{Q}\bm{K}^{T}}{\sqrt{d}})+\bm{X}_{mask}\Big)\bm{V},\\
			&\bm{O}=FFN(\bm{H}),
		\end{aligned}
	\end{equation}
	where $  \bm{Q}, \bm{K} $ and $\bm{V} \in  \mathbb{R}^{L\times d}$ are the  queries, keys and values of self-attention, for simplicity $ L= N+(\frac{T}{2}-1) \cdot \frac{H}{32} \cdot \frac{W}{32}$, $ \bm{W}_{q}, \bm{W}_{k}, \bm{W}_{v} \in  \mathbb{R}^{d\times d}$ are trainable parameter matrices, $\bm{H}$ is the hidden states, $ \bm{X}_{mask} $ is a token mask matrix, $ \bm{O} $ is the output of the layer, $ FFN(\cdot) $ is a feed-forward  sub-layer. While traditional self-attention mechanisms can directly capture the dependencies between input tokens, query and key are controlled by only two learnable matrices, missing the opportunity to exploit the shallow textual information, formally,
	\begin{equation}
		\label{eq_triss}
		\bm{Q}\bm{K}^{T}[i][j] =  \bm{I}[i](\bm{W}_{q}\bm{W}_{k}^{T})\bm{I}[j]^{T}.
	\end{equation}
	
	To solve this problem, we propose to add the output of the previous layers $ \bar{\bm{O}} $ as shallow textual  information to the $\bm{Q},\bm{K}$ calculation,
	\begin{equation}
		\label{eq_qk}
		\begin{aligned}
			& \hat{\bm{Q}} = (1- \lambda_{q})\bm{Q}+\lambda_{q} \bar{\bm{O}} \bm{W}_{oq},\\
			&\hat{\bm{K}} = (1- \lambda_{k})\bm{K}+\lambda_{k} \bar{\bm{O}} \bm{W}_{ok}	,\\
			&\lambda_{q} = sigmoid(\bm{Q}\bm{w}_{q}+\bar{\bm{O}} \bm{w}_{oq}),\\
			&\lambda_{k} = sigmoid(\bm{K}\bm{w}_{k}+\bar{\bm{O}} \bm{w}_{ok}),\\
			&\bar{\bm{O}} = mean(\bm{O}_{1},\bm{O}_{2}, \cdots,\bm{O}_{z-1}),	
		\end{aligned}
	\end{equation}
	where $ \bm{W}_{oq}, \bm{W}_{ok} \in \mathbb{R}^{d\times d} $ are trainable parameter matrices, $ \bm{w}_{q},\bm{w}_{k}, \bm{w}_{oq}, \bm{w}_{ok} \in \mathbb{R}^{d\times 1} $ are trainable parameter vectors, $ z $ is the order number of the current layer. Correspondingly, the output  is constructed based on shallow textual   information,
	\begin{equation}
		\label{eq_ourtr}
		\begin{aligned}	
			&\hat{\bm{H}}=\Big(softmax(\frac{\hat{\bm{Q}}\hat{\bm{K}}^{T}}{\sqrt{d}})+\bm{X}_{mask}\Big)\bm{V},\\
			&\hat{\bm{O}}=FFN(\hat{\bm{H}}).
		\end{aligned}
	\end{equation}
	
	Following \cite{Li_2020_Oscar,Hu_2021_VIVO}, we take the first $ N $ tokens of $ \hat{\bm{O}}_{Z} $ as the representation of the sentence, where $ \hat{\bm{O}}_{Z} $ is the output of the last layer of the decoder.
	
	\subsection{Training} 
	We train EVC-MF in an end-to-end manner and employ  Masked Language Modeling \cite{Devlin_2019_Bert} on our decoder. Specifically, we randomly mask out  a certain percentage words of the ground-truth by substituting them with \textit{[MASK]}. Subsequently, we utilize  the relevant output  of EVC-MF for classification to predict words. We adopt the standard cross-entropy (CE) loss to train EVC-MF, the loss
	for a single pair $ (\bm{V}, \bm{\mathcal{S}}) $ is,
	\begin{equation}
		\label{eq_loss}
		\mathcal{L} =  \sum_{\bm{w}_{n}\in \mathop{\bm{\mathcal{S}}}\limits^{ma}} \log P(\bm{w}_{n}|\mathop{\bm{\mathcal{S}}}\limits^{re},\bm{V} ) ,
	\end{equation}
	where  $ \bm{\mathcal{S}} = \{\bm{w}_{1},\bm{w}_{2},\cdots,\bm{w}_{N}\}$ is the ground-truth, $ \mathop{\bm{\mathcal{S}}}\limits^{ma} $ denotes  the set of masked words,  $ \mathop{\bm{\mathcal{S}}}\limits^{re} $  represents the set of remaining words.

	\subsection{Inference} 
	During inference, we generate the caption in an auto-regressive manner.  Concretely, 
	we initialize EVC-MF with a start token \textit{ [CLS] } and a \textit{[MASK]} token; then sample a word from the vocabulary based on the likelihood output. Subsequently, we replace the \textit{[MASK]} token in the previous input sequence with the sampled word and append a new \textit{[MASK]} for predicting the next word. The generation process terminates until the end token \textit{[EOS]} is generated or the maximum output length is reached.
	
	\section{Experiments}\label{section:experiment}
	\label{sec:experiments}
	In this section, we first compare EVC-MF with several  state-of-the-art methods for video captioning on two widely-used benchmark datasets, \textit{i.e.}  MSR-VTT and MSVD. Subsequently, to further illustrate the effectiveness of  EVC-MF, we conduct extensive ablation experiments, hyper-parametric analysis, and qualitative analysis.
	
	\subsection{Datasets}
	\textbf{MSVD} (Microsoft Video Description Corpus ) \cite{David_2011_MSVD} comprises a collection  of 1,970 YouTube  video clips  covering various topics including but not limited to baking, animals and landscapes, \textit{etc}. Each video clip lasts 9 to 10 seconds and focuses on a single activity. On average, there are approximately 42 ground-truth descriptions associated with each video clip, resulting in a total of around 8,000 English video-caption pairs. Following \cite{David_2011_MSVD}, we divide MSVD into training set, validation set and test set in the proportion of 60\%, 5\% and 35\%. 
	
	\textbf{MSR-VTT} (Microsoft Research Video to Text)  \cite{Xu_2016_MSRVTT} consists of 10,000 open domain video clips, encompassing a total of  200,000 video-description pairs. These videos cover diverse topics such as foods, movies, animals, landscapes, \textit{etc}. Furthermore, MSR-VTT also provides category tags and audio information for each video clip. Following the common settings, we split the dataset into a training set, a validation set and a test set consisting of 6,513, 497, 2,990 video clips, respectively.
	\begin{table*}[t]
		\small
		\center
		\begin{center}
			\caption{Performances of EVC-MF and other state-of-the-art methods on the MSVD and MSR-VTT datasets.}
			\resizebox{0.66\textwidth}{!}{
				\label{tab_comp}
				\begin{tabular}{c|c|llll|llll}
					\toprule[1pt]
					\multirow{2}*{Method} & \multirow{2}*{Features} & \multicolumn{4}{c|}{MSVD} & \multicolumn{4}{c}{MSR-VTT}\\
					&& B-4 & M & R & C & B-4 & M & R &C\\
					\hline
					\noalign{\smallskip}
					OSTG (2020) \cite{Zhang_2020_OSTC} & R200+MR & 57.5 & 36.8 & -& 92.1& 41.9 & 28.6 & - &48.2\\
					OpenBook (2021) \cite{Zhang_2021_OBVC}  & IRV2+C+T & - & - & -& - & 42.8 & 29.3 & 61.7 & 52.9\\
					TTA (2021) \cite{Tu_2021_TTA} & R152+C+MR  & 51.8 & 35.5 &72.4 & 87.7 & 41.4 & 27.7 & 61.1 & 46.7\\
					MGRMP (2021) \cite{Chen_2021_MGR} & IRV2+RN  & 55.8 & 36.9 &74.5 &98.5 & 41.7 & 28.9 & 62.1 & 51.4\\
					TVRD (2022) \cite{Wu_2022_TVRD} & IRV2+C+FR  & 50.5 & 34.5 &71.7 &84.3 & 43.0& 28.7 & 62.2 & 51.8\\
					vc-HRNAT (2022) \cite{Gao_2022_HRNAT} & IRV2+I  & 55.7 & 36.8 &74.1 & 98.1 & 42.1 & 28.0 & 61.6 & 48.2\\
					HMN (2022) \cite{Ye_2022_HMN}& IRV2+C+FR  & 59.2 & 37.7 &75.1 & 104.0 & 43.5 & 29.0 & 62.7 & 51.5\\
					HTG+HMG (2023) \cite{Tu_2023_HTG}& R+C  &52.7 & 35.2 &72.8 &91.4 & 42.1 & 28.4 & 61.6 & 48.9\\
					VTAR (2023) \cite{Shi_2023_VTAR}& IRV2+RN+T  &- & -&- &- & \underline{44.4} & \underline{30.0} & \underline{63.3} & \underline{56.2}\\
					\hline
					\noalign{\smallskip}
					XlanV (2020) \cite{Huang_2020_XlanV} & R152+I & - & - & -& - & 41.2 & 28.6 & 61.5 & 54.2\\
					SMAN (2022) \cite{Zheng_2022_SMAN} & IRV2+C+FR & 50.2 & 35.0 & 71.3& 87.7 & 41.3 & 28.7 & 62.1 & 53.8\\
					SHAN (2022) \cite{Deng_2022_SHAN} & IRV2+I & 50.9 & 35.1 & 72.4& 94.5 & 40.3 & 28.8 & 61.2 & 54.1\\
					CMG (2022) \cite{Wang_2022_CMG} & IRV2+C+a+c & \underline{59.5} & 38.8 & \underline{76.2}& 107.3 & 43.7 & 29.4 & 62.8 & 55.9\\
					\hline
					\noalign{\smallskip}
					SBAT (2020) \cite{Jin_2020_SBAT}&IRV2+I &53.1 & 35.3 & 72.3& 89.5 & 42.9 & 28.9 & 61.5 & 51.6\\
					STG-KD (2020)\cite{Pan_2020_STG} & R101+I+FR & 52.2 & 36.9 & 73.9& 93.0 & 40.5 & 28.3 & 60.9 & 47.1\\
					O2NA (2021) \cite{Liu_2021_O2NA} & R101+RN & 55.4 &37.4 &74.5 & 96.4 & 41.6 & 28.5 &62.4 & 51.1\\
					NACF (2021) \cite{Yang_2021_NA} & R101+RN+c  & 55.6 & 36.2 &- & 96.3 & 42.0 & 28.7 & - & 51.4\\
					LSRT (2022)\cite{Li_2022_LSRT} & IRV2+I+FR & 55.6 & 37.1 & 73.5& 98.5 & 42.6 & 28.3 & 61.0 & 49.5\\
					SWINBERT (2022)\cite{Lin_2022_swinbert} & VS & 56.7 & \underline{40.1} &\underline{76.2} & \underline{112.6} & 42.7 & \textbf{30.4} & 61.7 & 54.1\\
					\hline
					\noalign{\smallskip}
					EVC-MF  & VS  & \textbf{62.8} &\textbf{ 41.5} & \textbf{79.0} & \textbf{123.4} & \textbf{45.1} & \textbf{30.4} &\textbf{63.6} & \textbf{57.1}\\
					\toprule[1pt]
			\end{tabular}}
		\end{center}
	\end{table*}
	\subsection{Evaluation Metrics}
	In our experiments, we employ  four widely used metrics for quantitative evaluation: BLEU-4 \cite{Papineni_2002_BLEU_4}, METEOR \cite{Michael_2014_METEOR}, ROUGE-L \cite{Lin_2004_rouge} and CIDEr \cite{Vedantam_2015_CIDEr}. These metrics facilitate the evaluation of the quality of candidate sentences from various perspectives.
	
	 \textbf{BLEU} reflects the consistency between the candidate sentences and the ground-truth  sentences by calculating  their overlap in terms of $ n $-grams. Assuming  that the lengths of ground-truth sentence and candidate sentence are $ r $ and $ c $, respectively. The score of BELU is defined as,
		\begin{equation}
		\label{eq_belu}
		\begin{aligned}	
			&\text{BELU-N}=BP \cdot \exp\Big( \sum_{n=1}^{N}  w_{n}  \log p_{n}  \Big), \\
			&BP=	\begin{cases} 
				1,  & \mbox{if } c>r,\\
				\exp(1-r/c), & \mbox{if } c\leq r,
			\end{cases}
		\end{aligned}
	\end{equation}
	where $ BP $ is a brevity penalty, $ p_{n} $ is the $ n $-gram precision, $ w_{n} $ is a positive weight usually taken as $ 1/n $.
	
    \textbf{METEOR} takes into account the accuracy and recall rate of the entire corpus. First unigram precision $ P_{u} $ and unigram recall $ R_{u} $  are computed as the ratio of the number of unigrams in candidate sentences  that are mapped to unigrams in the ground-truth  sentences. Then, the METEOR score for the given alignment is computed as follows:
    \begin{equation}
    	\label{eq_meteor}
    	\begin{aligned}	
    		&\text{METEOR}=(1-Pen)F_{mean} , \\
    		&F_{mean} = \frac{(\alpha_{m}^{2}+1)P_{u}R_{u}}{R_{u}+\alpha_{m}^{2}P_{u}}, \\
    		&Pen=\gamma_{m}(\frac{ch}{um})^{\theta_{m}},	
    	\end{aligned}
    \end{equation}
   where  $ F_{mean} $ is a harmonic mean; $ Pen $ is a fluency  penalty; $ ch  $  and $ um $  denotes the number of chunks and  unigram on the given alignment; $ \alpha_{m}, \gamma_{m}$ and $ \theta_{m} $ are usually set to $ \alpha_{m}=3$ , $ \gamma_{m}=0.5$ and $ \theta_{m} =3$, respectively.
    
    \textbf{Rouge-L} calculates the length of the longest common subsequence between the candidate sentences and  ground-truth sentences. The score of BELU is defined as,
    \begin{equation}
    	\label{eq_Rouge}
    		\text{Rouge-L}= \frac{(\alpha_{r}^{2}+1)P_{r}R_{r}}{R_{r}+\alpha_{r}^{2}P_{r}}, 
    \end{equation}
     where $ \alpha_{r} $ is a hyper-parameter usually takes a large value, $ P_{r}  $ and $ R_{r} $ denote the recall and accuracy of candidate and  ground-truth sentences based on the longest common subsequence, respectively. 
    
    \textbf{ CIDEr} integrates  BLEU with a vector space model to evaluate whether the model captures critical information. Firstly, the number of times an n-gram $ w_{k} $ occurs in a ground-truth sentence $ \bm{\mathcal{S}}_{ij} $ (the $ j $-th ground-truth sentence of the $ i $-th video) or  candidate sentence  $ \hat{\bm{\mathcal{S}}}_{i} $ is denoted by $ h_{k}(\bm{\mathcal{S}}_{ij}) $ or $ h_{k}(\hat{\bm{\mathcal{S}}}_{i}) $. The TF-IDF weighting $ g_{k}(\bm{\mathcal{S}}_{ij}) $ for each n-gram $ w_{k} $ is defined as:
     \begin{equation}
    	\label{eq_gk}
    	g_{k}(\bm{\mathcal{S}}_{ij})=  \frac{h_{k}(\bm{\mathcal{S}}_{ij})}{\sum_{w_{l}\in \Omega}} \log \Big(\frac{|V|}{\sum_{V_{p}\in V} min(1, \sum_q h_{k}(\bm{\mathcal{S}}_{pq}))}\Big),
    \end{equation}
    where $ \Omega $ is the vocabulary of all n-grams and $ V $ is the set of all videos in the dataset. The score of CIDEr is defined as:
     \begin{equation}
    	\label{eq_CIDEr}
    	\begin{aligned}	
        	&\text{CIDEr}= \sum_{n=1}^{N} w_{n} \text{CIDEr}_{n},\\
        	&\text{CIDEr}_{n} = \frac{1}{m} \sum_{j} \frac{\bm{g}^{n}(\hat{\bm{\mathcal{S}}}_{i})\bm{g}^{n}(\bm{\mathcal{S}}_{ij})}{||\bm{g}^{n}(\hat{\bm{\mathcal{S}}}_{i})\|||\bm{g}^{n}(\bm{\mathcal{S}}_{ij})||}
    	\end{aligned}
    \end{equation}
    where $ \bm{g}^{n}(\hat{\bm{\mathcal{S}}}_{i}) $  is a vector formed by $ g_{k}(\hat{\bm{\mathcal{S}}}_{i}) $  corresponding to
    all n-grams of length n, similarly for $ \bm{g}^{n}(\bm{\mathcal{S}}_{ij}) $; $ m $ is the number of captions corresponding to the video; $ w_{n} $ is a positive weight usually taken as $ 1/n $.
    
    For all evaluation metrics, the better the quality of captions is, the higher the scores are. For convenience, in the rest of the paper, we use B-4, R, M and C denote  BLEU-4, ROUGE-L, METEOR and CIDEr, respectively.
	
	\subsection{Implementation Details}
	\textbf{Video Preprocessing.} We uniformly sample $ 32$ ($ T=32 $) raw frames for each video clip in both datasets. Subsequently, we resize the frames to $ 224 \times 224 $ ($ H=W=224 $) to fit the size of VidSwin. 
	
	\textbf{Feature Extractor.} The VidSwin is pre-trained on the Kinetics-600 dataset. To fine tune the feature extractor,  we set its learning rate  to be $ 0.05 $ times than that of other modules.
	
	\textbf{Encoder.} We define the grid size as $ 4\times 4 $ ($ g=4 $),  the area of  the smallest region as  $ 2g\times 2g$ ($ \Delta x=\Delta y=2 $). Furthermore, we set the hyper-parameter $ \delta $ to $ 0.3 $. Consequently, we obtain  $ 575 $ ($ |\mathcal{R}|=575 $) different regions.
	
	\textbf{Decoder.} Our decoder has $ 4 $ enhanced transformer layers ($ Z=4 $). And we set the  dimensionality of the hidden layer features to $ 768 $ ($ d=768 $). Furthermore, the vocabulary size amounts to $ 30,522 $. In the training phase, we set the maximum length of the sentences and the mask rate  to $ 50 $ and $ 0.5 $ , respectively. In the inference phase, we limit the maximum sentence length to $ 20 $. Additionally, we use beam search to generate final sentences, the beam size is set to $ 4 $.

	\textbf{Other details.} We use  Pytorch \cite{Paszke_2019_PyTorch}, Deep-Speed library \cite{Rasley_2020_DeepSpeed} to implement EVC-MF. During  the training phase, we utilize Adam algorithm with batch size of $ 6 $ and gradient accumulation steps of $ 4 $. The learning rates, warmup ratio, and weight decay for both MSR-VTT and MSVD are set to $ 4 \times 10^{-5} $, $ 0.1 $  and $ 0.05 $, respectively. The maximum epochs  is set to $ 50 $. Furthermore, all experiments are conducted on a single NVIDIA RTX3090 GPU with 24GB RAM and the OS of our server is Ubuntu16.04 with 328G RAM. 
	
	\subsection{Comparisons with State-of-the-Art Methods}
	In order to verify the effectiveness of EVC-MF for video captioning, we conduct a comprehensive evaluation against state-of-the-art methods. The results on MSVD and MSR-VTT are showed in Table \ref{tab_comp}. Following the conventional setting, we report all results as percentages (\%), with the highest and second-highest scores  shown in bold and underlined, respectively. Depending on decoder type and utilization of sequence optimization techniques, \textit{i.e.} reinforcement learning used, we separate previous approaches into three parts: 1) The models in the first part use RNN-based decoders without  sequence optimization, \textit{ e.g.} OpenBook, MGRMP, \textit{etc.}, 2) The models in the second part utilize  RNN-based decoder with sequence optimization, \textit{ e.g.} SMAN, CMG, \textit{etc.}, 3) The models  in the third part adopt  transformer-based decoder without sequence optimization, \textit{e.g.} STG-KD, SWINBERT, \textit{etc.}  
	
	For a fair comparison, we present  the best results of these methods on both MSVD and MSR-VTT  test sets. It is worth mentioning that, SWINBERT \cite{Lin_2022_swinbert} only reports  results on the validation  set in the original paper. To maintain fairness, we have reproduced it using  the code\footnote{https://github.com/microsoft/SwinBERT} published by the authors.
	
	Furthermore,  in Table \ref{tab_comp}, the abbreviated names IRV2, R*, RN, I, C, FR, MR, VS, T, a and c denotes Inception-ResNet-V2, ResNet*, 3D-ResNext-101, I3D, C3D, Fast-RCNN, Mask-RCNN, VidSwin, Pre-Retrieval Text,  audio information (MSR-VTT only) and category information (MSR-VTT only), respectively, where $ * \in \{101, 152,500\} $. In addition, "-" indicates the absence of results for this metric in the original paper.
	From this Table \ref{tab_comp}, we have the following observations:
	\begin{itemize}[leftmargin=*]
		\item On both datasets, EVC-MF achieves  the best results in terms of all widely-used metrics, especially on the  more in line with human judgment metric, CIDEr. For example, EVC-MF is 10.8\% and 0.9\% higher than the runner-up methods in terms of CIDEr on MSVD and MSR-VTT, respectively.
		\item The first and third parts of the table demonstrate that both RNN-based decoder and transformer-based decoder  exhibit superior performance. However, it is noteworthy that transformers can be trained in parallel, thereby offering convenience for end-to-end training. Consequently, we  choose the  transformer-based decoder to generate sentences.
		\item From the second part of the table,  it can be observed that  methods using sequence optimization, \textit{e.g.} SMAN, perform well  in terms of CIDEr. This can attributed to their utilization of CIDEr as the target for sequence optimization. Notably, EVC-MF  solely relies on cross-entropy loss optimization model; however, it even surpasses them with respect to all metrics. This further substantiates the efficacy of our proposed method.
		\item When using the same feature extractor, \textit{i.e.} VidSwin, EVC-MF surpasses  the recently  proposed SWINBERT \cite{Lin_2022_swinbert}, which is also an end-to-end model for video captioning. Furthermore, both end-to-end training approaches, \textit{i.e.} SWINBERT and EVC-MF, using raw video as input  exhibit significant advancements over alternative methods, especially on MSVD.
		\item CMG  exhibits suboptimal performance across most metrics on MSVD. This may be due to the fact that more additional information can provide   different perspectives on the understanding of the video. It is worth pointing out that getting features from different levels is also a way for EVC-MF to understand the video from different perspectives.
	\end{itemize}
	\subsection{Ablation Study}
	\begin{table}[t]
		\small
		\center
		\begin{center}
			\caption{Performance of EVC-MF on MSR-VTT with different add-on components.}
			\resizebox{0.40\textwidth}{!}{
				\label{tab_abl}
				\begin{tabular}{c|llll}
					\toprule[1pt]
					\multirow{2}*{Method} & \multicolumn{4}{c}{MSR-VTT}\\
					&  B-4 & M & R & C \\
					\hline
					\noalign{\smallskip}
					Baseline  &  42.4 & 28.3 & 61.8 & 52.1\\
					\hline
					\noalign{\smallskip}
					EVC-MF w/o ME and ET  &  43.4 & 29.3 & 62.5 & 53.4\\
					EVC-MF w/o  ET   &  44.3 & 29.4 & 63.0 & 56.2\\
					EVC-MF w/o  ME   &  43.8 & 29.3 & 62.8 & 55.5\\
					\hline
					\noalign{\smallskip}
					EVC-MF    &  \textbf{45.1} & \textbf{30.4} & \textbf{63.6} & \textbf{57.1}\\
					\toprule[1pt]
			\end{tabular}}
		\end{center}
	\end{table}
	To demonstrate the effectiveness of all the modules in  EVC-MF, we further conduct ablation experiments on MSR-VTT. For this purpose, we design a baseline model comprising an identical feature extractor to that of  EVC-MF, but only extracting features before the classifier, as well as a  traditional transformer-based decoder to generate captions. Subsequently, we further denote the modules used in EVC-MF as follows:
	\begin{itemize}[leftmargin=*]
		\item MF: the multi-scale features are employed  in the model;
		\item ME: the masked encoder is used in the model;
		\item ET: the enhanced transformer layer is utilized  in the model.
	\end{itemize}
	
	The results of the models with different modules on MSR-VTT are reported in Table \ref{tab_abl}. From this table, we can observe that:
	\begin{itemize}[leftmargin=*]
		\item EVC-MF with only MF (the second row) exhibits significant improvement. For example, there is an improvement of 1.0\% and 1.8\% on BLEU-4 and CIDEr. This observation underscores utility of incorporating shallow visual information for video comprehension.
		\item The contributions of ME and ET are also noteworthy, their absence leads to a decrease in the CIDEr metric of EVC-MF by 1.6\% and 0.9\%, respectively.
		\item In summary, when each sub-module is added, the results in terms of all  widely-used metrics are improved, which demonstrates the effectiveness of the sub-modules.
		
	\end{itemize}
	\begin{figure*}
		\centering
		\centerline{\includegraphics[width=18cm]{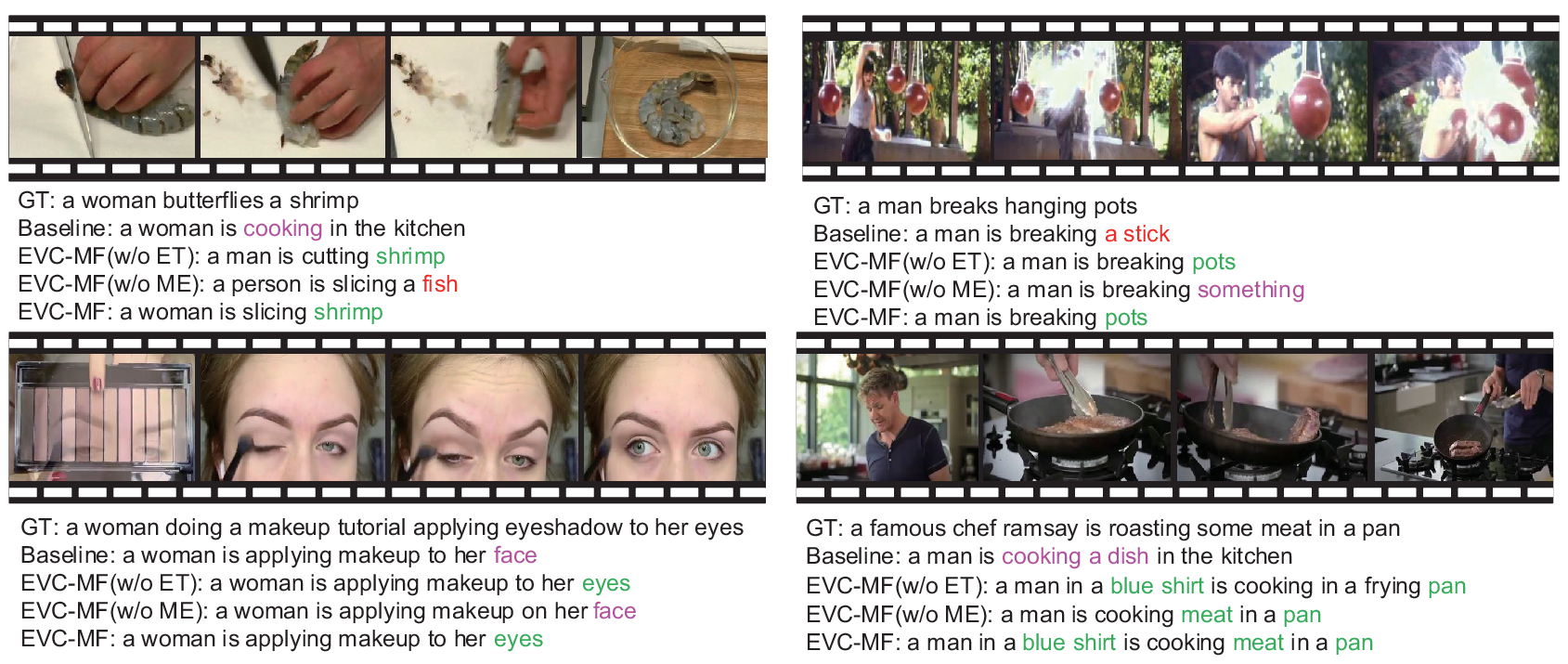}}
		\caption{Qualitative results on MSVD and MSR-VTT. The first row is from MSVD and the second is from MSR-VTT. Correct descriptions are marked in green, while wrong and inaccurate words are marked as red and purple respectively.}
		\label{fig_QR}
	\end{figure*}
	\subsection{Evaluation of hyper-parameters}
	\begin{table}[t]
		\small
		\center
		\begin{center}
			\caption{Performance of EVC-MF on MSR-VTT with different smallest region.}
			\resizebox{0.32\textwidth}{!}{
				\label{tab_ds}
				\begin{tabular}{c|llll}
					\toprule[1pt]
					\multirow{2}*{	$ \Delta x$ and $ \Delta y$}  & \multicolumn{4}{c}{MSR-VTT}\\
					&  B-4 & M & R & C \\
					\hline
					\noalign{\smallskip}
					1  &  44.6 & 30.0 & 63.1 & 56.3\\	
					2  &  \textbf{45.1} & 30.4 & \textbf{63.6} & \textbf{57.1}\\
					3   &  44.8 & \textbf{30.6} & 63.2 & 56.9\\
					4   &  45.0 & 29.5 & 62.0 & 55.0\\		
					\toprule[1pt]
			\end{tabular}}
		\end{center}
	\end{table}
	
	\textbf{$ \Delta x$ and $\Delta y$} of ME determine the area of the smallest region in masked encoder. To evaluate their effect on EVC-MF, we conduct experiments
	on MSR-VTT  with varying values for $ \Delta x$ and $\Delta y$. The results are
	summarized in Table \ref{tab_ds}. From Table \ref{tab_ds}, we have the following observations:
	\begin{itemize}[leftmargin=*]
		\item EVC-MF obtains relatively favorable  results on MSR-VTT when $ \Delta x=\Delta y=2$.
		\item Generally speaking, when the region is too small or too large, the performance of EVC-MF deteriorates slightly. One potential explanation for this issue lies in the fact that if $\Delta x$ and $\Delta y$ are excessively small, ME may not exert  a  sufficiently significant influence on the process; whereas if $\Delta x$ and $\Delta y$ are excessively  large, crucial  information might inadvertently be obscured.
	\end{itemize}
	\textbf{The threshold $ \delta $} of $ R_{i,j} $ of masked encoder determines the masked area. Consequently, it also significantly influences the performance  of EVC-MF. To investigate  this impact, we conduct experiments on MSR-VTT when $ \delta $ takes different values.  The corresponding  results are  presented  in \ref{tab_dl}.  From which, we have the following observations:
	\begin{itemize}[leftmargin=*]
		\item EVC-MF performs best when $ \delta =0.3$.
		\item Jointly considering the results in Table \ref{tab_ds} \& \ref{tab_dl}, it becomes evident that the performance of EVC-MF significantly deteriorates  when the average area of the sequence of masked regions $ \tilde{\mathcal{R}} $ is excessively large. One of the main reasons is that a large amount of information is lost, when most of the feature maps of the sequence are masked by larger areas. 
		
	\end{itemize}
	\begin{table}[t]
		\small
		\center
		\begin{center}
			\caption{Performance of EVC-MF on MSR-VTT with different values of  $ \delta $.}
			\resizebox{0.26\textwidth}{!}{
				\label{tab_dl}
				\begin{tabular}{c|llll}
					\toprule[1pt]
					\multirow{2}*{$ \delta $ }  & \multicolumn{4}{c}{MSR-VTT}\\
					&  B-4 & M & R & C \\
					\hline
					\noalign{\smallskip}
					0.0  &  43.8 & 29.3 & 62.8 & 55.5\\	
					0.3  &  \textbf{45.1} & \textbf{30.4} & \textbf{63.6} & \textbf{57.1}\\
					0.5   &  43.7 & 29.0 & 62.3 & 55.2\\
					1.0   &  43.2 & 29.4 & 62.6 & 54.6\\		
					\toprule[1pt]
			\end{tabular}}
		\end{center}
	\end{table}
	
	\subsection{Qualitative Results}
	To intuitively analyze the effectiveness of EVC-MF, we present some illustrative  cases from MSVD and MSR-VTT in Fig. \ref{fig_QR}. Among them, GT represents the ground-truth, while the other settings remain consistent with those in Table \ref{tab_abl}.
	As depicted in Fig. \ref{fig_QR}, while baselines, EVC-MF(w/o ET) and EVC-MF(w/o ME)  mistakenly interpret  the video contents, EVC-MF  accurately captures relevant words  and generates more precise and comprehensive captions. Specifically, the example in the top left, the baseline model only generates a generalized word "cooking", EVC-MF(w/o ME) even generates an error description "fish". In contrast, captions generated by EVC-MF are apparently more precise. Similar situations  occurs in other examples as well. Surprisingly, in the example at the bottom left, both  EVC-MF and EVC-MF(w/o ME) generate a more detailed content description of "blue shirt", which is present in the video but absent from the ground-truth. This phenomenon further demonstrates  that masked encoder can facilitate learning of more detailed and useful information.
	\section{Conclusion }\label{section:conclusion}
	In this paper,  we propose a novel end-to-end encoder-decoder-based network (EVC-MF) for video captioning, comprising a feature extractor, a masked encoder, and an enhanced transformer-based decoder. Specifically, to ensure updatable parameters of the feature extractor and optimize the utilization of shallow visual information, the feature extractor takes the original frame as input and extracts multi-scale visual features to the encoder. Then, to learn more valuable  details, extract meaningful insights, and reduce unnecessary redundancy, we propose a  masked encoder. Finally, to fully utilize visual and text information, we  develop an enhanced transformer-based decoder. Furthermore, we conducted extensive experiments on MSVD and MSR-VTT to demonstrate the effectiveness of  EVC-MF and its sub-modules. Although EVC-MF achieves better performance, it still lacks in controllability and interpretability. Thus, in the future, we will work on improving it in these two aspects.

	\bibliographystyle{IEEEtran}
	\bibliography{sigproc}

\end{document}